# Machine Learning-Based Classification of Jhana Advanced Concentrative Absorption Meditation Using 7 Tesla Functional Magnetic Resonance Imaging

Puneet Kumar[1], Winson F.Z. Yang[2,3], Alakhsimar Singh[4], Xiaobai Li[5], Matthew D. Sacchet[2,3*]

**Affiliations**

[1]School of AI and Data Engineering, Indian Institute of Technology Ropar 140001, India

[2]Meditation Research Program, Department of Psychiatry, Massachusetts General Hospital, Harvard Medical School, Boston, MA 02129, USA

[3]Athinoula A. Martinos Center for Biomedical Imaging, Department of Radiology, Massachusetts General Hospital, Harvard Medical School, Boston, MA 02129, USA

[4]Computer Science and Engineering Department, National Institute of Technology Jalandhar 144008, India

[5]State Key Laboratory of Blockchain and Data Security, Zhejiang University, Hangzhou, 310027, China.

[*]**Corresponding author:**

Meditation Research Program, Department of Psychiatry, Massachusetts General Hospital, Harvard Medical School, Boston, MA 02129, USA.

Email: meditationadministration@mgh.harvard.edu

## Abstract

**Introduction:** Jhana advanced concentrative absorption meditation (ACAM-J) is related to profound changes in consciousness and cognitive processing, making the study of its neural correlates vital for insights into consciousness and well-being. Neuroimaging work on ACAM-J has so far relied on univariate, group-level contrasts, leaving open whether these states carry distributed neural signatures that can be decoded from individual scans. This study evaluates whether functional MRI-derived regional homogeneity (ReHo) can be used to classify ACAM-J using machine-learning approaches.

**Methods**: We analysed 7T fMRI data from 20 advanced meditators in the group study who progressed through their standard ACAM-J sequence and two matched non-meditative control tasks, together with intensive single-case data from one separate case-study participant held out for final evaluation. ReHo maps were computed for every fMRI segment and parcellated into 498 regions of interest spanning cortex, subcortex, brainstem, and cerebellum. Within subject-wise stratified cross-validation, feature ranking, recursive feature elimination, and class balancing were applied to training data only; six classifier families were fitted, and the three best-performing models per contrast were combined by probability averaging.

**Results**: Across 19 binary comparisons, the ensemble reached an overall average accuracy of 65.87% before and 66.82% after variance associated with reported phenomenology was regressed out, with an overall average Cohen's κ of 0.2443 before residualization. Discrimination was strongest for the most widely separated absorption states (ACAM-J1 vs ACAM-J6, 74.33% accuracy, κ = 0.5158), while adjacent states were harder to separate. Feature-importance analysis indicated that prefrontal and anterior cingulate areas contributed most to model decisions.

**Conclusion:** Together, these findings provide evidence that ReHo patterns measured at 7T carry information that distinguishes ACAM-J from cognitively active control states and, to a more limited degree, distinguishes ACAM-J states from one another. They support the feasibility of multivariate decoding of advanced meditation and may inform future work on the mechanisms and neuromodulation of these states.

## Key Points:

1) 7T fMRI-derived regional homogeneity, decoded with machine learning, distinguishes jhana advanced concentrative absorption meditation (ACAM-J) from matched non-meditative control tasks, with weaker but above-chance discrimination among ACAM-J states themselves.

2) The most consistently contributing regions, including subcortical and brainstem structures difficult to characterize at lower field strengths, implicate attention, salience, and arousal-related systems in advanced absorption.

3) The study provides a structured, fully documented decoding workflow for small-sample, high-dimensional neuroimaging, including subject-wise cross-validation, leakage-controlled feature selection, and ensemble modelling, validated on a held-out intensively sampled case and reusable for other rare cognitive states.

## 1 Introduction

With advanced meditation training, human consciousness can reorganize into highly structured, non-ordinary modes of awareness (Sacchet et al., 2024). One such training, described in contemplative traditions, involves deep absorption in which attention becomes stable, sensory processing and inner narration diminish, and experiences of inner joy develop. Unlike the heterogeneous effects of more basic forms of meditation (Kang & Whittingham, 2010;

Tang et al., 2015), these advanced meditation states are structured, reproducible, and with distinct and reportable phenomenology (Galante et al., 2023; Ganesan et al., 2024; Potash et al., 2025; Sparby & Sacchet, 2024; Wright et al., 2023; Yang et al., 2024a). These experiences remain almost entirely unstudied in modern science, despite their significance in contemplative traditions (Yang et al., 2024b). Here, we view these states as alternate attentional configurations that provide a complementary reference point for studying peak human experiences, sustained attention, equanimity, and well-being (Sacchet et al., 2024; Sparby & Sacchet, 2025a; Sparby & Sacchet, 2025b).

Among the most systematically described and replicable frameworks for advanced meditation in contemplative traditions is jhana, a sequence of eight absorption states preserved in Buddhism for over two thousand years (Collins, 2025; Dhammadharo, 2017; Gunaratana, 1988). In this work, we refer to the practice in scientific terms as jhana advanced concentrative absorption meditation (ACAM-J) to differentiate the neurocognitive phenomena under study from doctrinal or interpretive accounts (Yang et al., 2024b). This terminology allows a precise scientific treatment grounded in first-person phenomenology while avoiding long-standing definitional debates surrounding jhana. ACAM-J comprises eight progressive states of advanced meditation that unfold through increasingly refined absorption and attentional stability (Sparby & Sacchet, 2025b). Early ACAM-J are marked by pronounced positive affect, including feelings of joy, bliss, and energized yet stable attention. In contrast, later ACAM-J are characterized by a profound sense of formlessness, accompanied by diminished sensory content and reduced self-referential thought. Awareness in these stages becomes expansive, boundless, and subtle, often described as a spacious void or a mode of consciousness that transcends ordinary perception. Each ACAM-J is phenomenologically distinct and can be reliably reported by experienced practitioners (Sparby & Sacchet, 2024). Despite its scientific importance to the science of consciousness and clinical potential (Abellaneda-Pérez et al., 2024; Potash et al., 2024; Yang et al., 2024b), systematic neuroscientific investigation of ACAM-J remains limited.

Recent neuroscientific investigations of ACAM-J have mapped its neural dynamics using high-resolution EEG and fMRI, revealing distinct brain states associated with concentrative absorption (Chowdhury et al., 2025; Demir et al., 2025; Treves et al., 2025). However, these studies rely on univariate analyses, which can detect group-level differences but not the high-dimensional "neural fingerprint" distinguishing brain states that machine learning (ML) offers (Jordan & Mitchell, 2015). ML applications to standard meditation practices have uncovered insights into neural mechanisms and associated cognitive changes (Hsu et al., 2024; S. Singh et al., 2022). Extending these approaches to advanced meditation would allow tracking of neural patterns during absorption, support the discovery of candidate physiological markers that may relate to meditative depth and clinical outcomes, and provide data-driven criteria for distinguishing ACAM-J (Ganesan et al., 2024; Hsu et al., 2024; Potash et al., 2024; Webb et al., 2022).

In this study, we ask whether ACAM-J states can be distinguished from cognitively active control conditions, and whether progressively different ACAM-J states can be decoded from distributed ReHo patterns measured with 7T fMRI. We address this question with a decoding pipeline built from established components, namely recursive feature elimination, class balancing for severely imbalanced designs, subject-wise stratified cross-validation, and ensemble averaging, applied to 498 ROI-level ReHo features. The resulting classification provides quantitative evidence that ACAM-J can be discriminated from matched non-meditative control conditions, while pairwise decoding across ACAM-J examines the extent to which progressively different absorption states can be distinguished. Feature-importance and ROI overlap analyses identify brain regions that most consistently contribute to classification, including subcortical and brainstem structures accessible with 7T fMRI. We further examine how classification behaves after accounting for variance associated with structured first-person reports and evaluate the fitted models on an intensively sampled, held-out single-case dataset, providing a cross-participant test of generalization. Together, these analyses provide a structured and interpretable decoding workflow for small-sample, high-dimensional neuroimaging and characterize distributed neural patterns associated with advanced absorption.

## 2 Methods

**Fig. 1** presents an overview of the proposed ACAM-J ML classification pipeline, which consists of five sequential modules. Module I covers data collection and preprocessing, including data acquisition, standard preprocessing steps, and ReHo calculation. Module II focuses on feature engineering and selection by applying SMOTE to balance the dataset and identifying relevant features based on prior studies (Yang et al., 2024a) and Recursive Feature Elimination (RFE). Module III implements the machine learning procedures integrating multiple algorithms with five-fold subject-wise stratified cross-validation and ensemble learning. Module IV evaluates model performance using case-study data, and Module V analyzes feature significance by generating brain maps that highlight the most critical regions for ACAM-J classification. The following sections describe each module in detail.

### 2.1 Participants and preprocessing

#### 2.1.1 Participants

The study combined data from two complementary cohorts, a case study participant from our previous work (Yang et al., 2024a), and group-level cohort data. The case study participant was a white American male adept meditator who was 52 years old at the time of the study. The participant was a long-term meditation teacher with over 25 years of meditation experience. Based on an estimated one to two hours of daily practice since the start of practice, and approximately a cumulative one year of retreat at 14 hours per day, we estimate a total practice amount of at least 20,000 hours. 20 participants (19 males, 1 female) from well-established meditation communities, all of whom demonstrated advanced expertise in ACAM-J meditation practices, were recruited for the group-cohort study. On average, the age range in this sample was 26-67 years old (mean age = 43.95 ± 15.10 years). On average, participants had 17.40 ± 11.59 years of meditation experience and reported an estimated total of 20500.40 ± 17288.98 hours of dedicated meditation practice. The study received approval from the Mass General Brigham Institutional Review Board (IRB), and the participant provided informed consent.

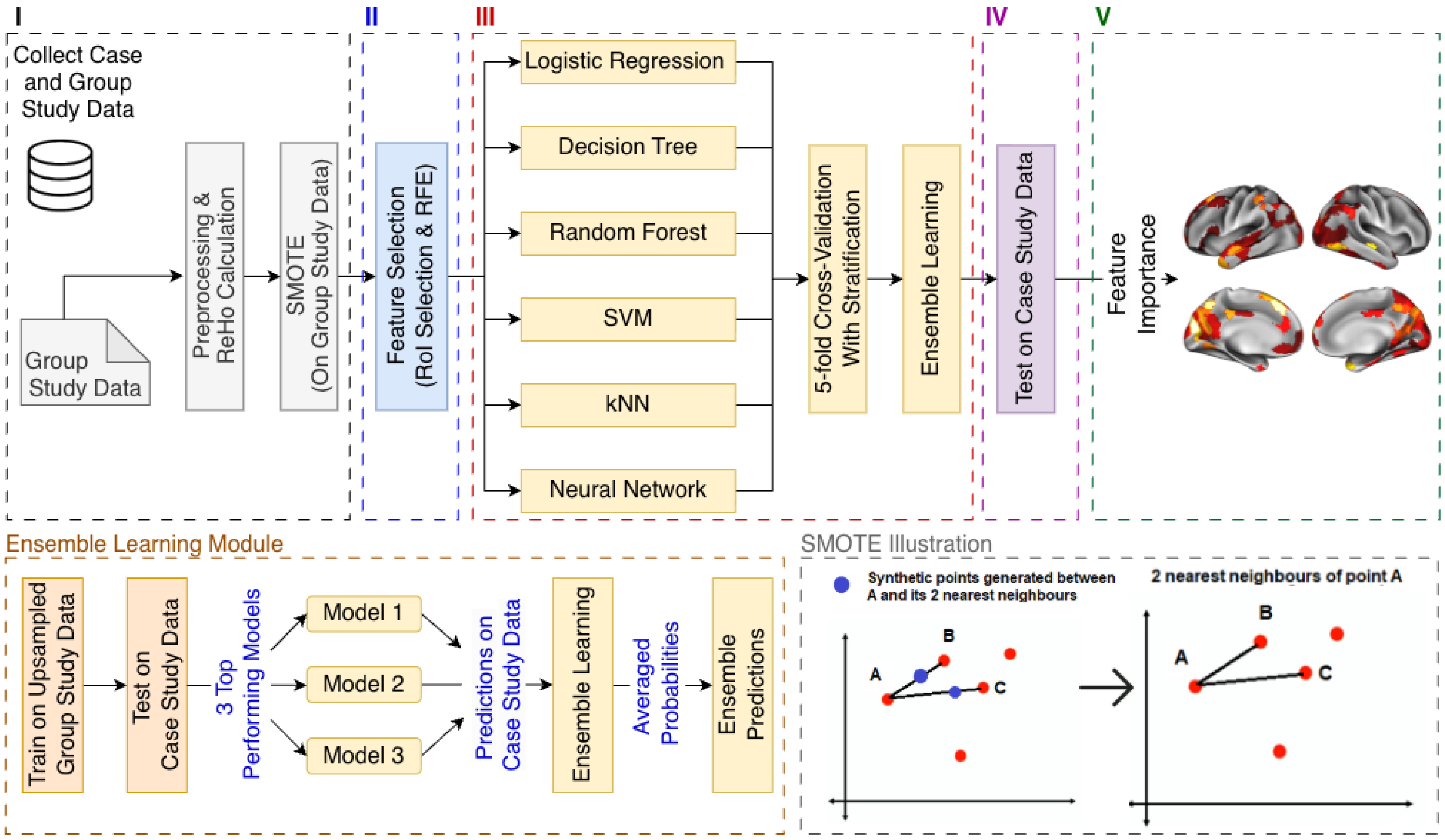

*Fig. 1:* ***Machine Learning Pipeline for Classifying ACAM-J, consisting of five modules****: I. Data Collection and Preprocessing; II. Feature Engineering and Selection; III. Machine Learning; IV. Model Testing and Evaluation; V. Feature Significance Analysis. Ensemble Learning and SMOTE modules are elaborated separately. The figure illustrates raw fMRI acquisition and ReHo computation, ROI balancing via SMOTE and pruning via RFE ensemble classifier training under 5-fold subject-wise stratified cross-validation, a two-step group- and case-study evaluation scheme, and derivation and aggregation of normalized feature-importance metrics across models. Implementation specifics are provided in the Supplementary Section.*

### 2.1.2 Ethics Statement

This study was approved by the Mass General Brigham Institutional Review Board (IRB; protocol no. 2019P003902). All participants provided written informed consent prior to participation. The study was conducted in accordance with institutional requirements and applicable regulations.

### 2.1.3 Experimental design

Both case study and group-level data were collected following the protocol outlined in our previous study (Yang et al., 2024a). Below is a summary of the data collected from the cohort study.

#### *2.1.3.1 Advanced concentrative absorption meditation (ACAM-J)*

Participants followed their standard ACAM-J sequence with closed eyes, progressing from access concentration through ACAM-J1 to ACAM-J8, concluding with a post-ACAM-J8 state referred to as afterglow. We requested participants to stay within any ACAM-J for at least 3 minutes for reliable computation of fMRI metrics (Wilson et al., 2015). For participants with ACAM-J duration of less than 3 minutes, we conducted another session of ACAM-J. The participants marked transitions from access concentration to ACAM-J8 with a button press. Note that 2 participants, including one from the case study, did not indicate transitions from ACAM-J6 to ACAM-J7 nor from ACAM-J7 to ACAM-J8, as it would disrupt the natural flow of ACAM-J practice for them. Please see Yang et al. (2025) for additional details.

#### *2.1.3.2 Non-meditative control tasks*

Two non-meditative control conditions were developed for comparison with meditation states, aimed at engaging the participant's mind in non-meditative, everyday cognitions while avoiding the induction of meditative states. Resting-state control task was not used due to concerns that experienced meditators might enter meditative states during rest, which may complicate interpretations of results (Tang et al., 2012). These control conditions consider the phenomenology of meditative states and the nature of the experimental paradigm to ensure greater consistency and reliability in the data (Yang et al., 2024a). The two control conditions involved a memory task, where the participant recalled events from the past one week, and a counting task, where the participant mentally counted down from 10,000 in decrements of 5. For most participants, one 8-minute run of each control task was collected. A subset of participants completed a second set of control runs because they were also involved in a separate, unrelated case-study project that required two sets of control conditions.

#### *2.1.3.3 Phenomenology*

To complement objective neuroimaging data, this study design also used a first-person phenomenological approach, in which the subject assessed the mental and physiological processes relevant to the experience of ACAM-J during meditation. Following a complete ACAM-J run (ACAM-J1 to ACAM-J8), the participant assigned a rating from 1 to 10 for phenomenological items including: (1) stability of attention, ranging from poor to excellent stability; (2) width of attention, varying from a narrow scope like a laser to very wide like a fisheye lens; (3) intensity of ACAM-J, indicating how intense a specific ACAM-J quality was; (4) early phenomenology of sights, sounds, physical sensations, and narrative thought stream, denoted by their presence during the form ACAM-J (ACAM-J1-J4); and (5) late phenomenology of sights, sounds, physical sensations, and narrative thought stream, characterized by their presence during the formless ACAM-J (ACAM-J5-J8). Higher rating values indicated a more robust experience within the given item. Stability of attention, width of attention, and intensity of ACAM-J were rated for each ACAM-J, while phenomenology of sights, sounds, physical sensations, and narrative thought stream were rated separately for early (ACAM-J1-J4) and late (ACAM-J5-J8) aggregate states.

### 2.1.4 Neuroimaging acquisition and preprocessing

Following the detailed protocols in Yang et al. (2024a), neuroimaging scans were obtained utilizing a 7T magnetic resonance (MR) scanner (SIEMENS MAGNETOM Terra) equipped with a 32-channel head coil. Functional imaging utilized a single-shot two-dimensional echo planar imaging sequence with T2∗-weighted BOLD-sensitive MRI. Whole-brain T1-weighted structural images were also acquired with the same parameters as the case study (Yang et al., 2024a). Physiological signals, including heart rate (measured using pulse oximetry) and respiration (recorded with breathing bellows), were continuously monitored throughout the scanning session.

Preprocessing of fMRI followed standard preprocessing steps and encompassed the following: (1) de-spiking, (2) RETROspective Image CORrection (RETROICOR), (3) slice time correction, (4) distortion correction, (5) motion correction, (6) registration of the anatomical dataset (T1w) to a standard MNI template, (7) scrubbing and removal of any volume with motion >0.3 mm and more than 5% outlier voxels, (8) regression of eroded cerebrospinal fluid (CSF) mask time course and motion parameters (3 translations, 3 rotations) per run, and finally band-pass filtering (0.01 to 0.1 Hz). Each fMRI run was then partitioned into distinct segments corresponding to different ACAM-J.

### 2.1.5 Regional homogeneity (ReHo) calculation

ReHo is a measure of similarity in temporal activation pattern of a voxel and its nearby voxels (Han et al., 2019; Yang et al., 2024a; Zang et al., 2004). This measure of local functional connectivity within brain regions is a close derivative of underlying brain activity. A higher ReHo value indicates stronger synchronization of local brain regions, indicating greater functional connectivity in that region and thus an index of greater brain activity. In this study, we defined a cluster size of 27 and calculated the ReHo values for each fMRI segment, standardized the ReHo values before smoothing the standardized ReHo maps using a 2 mm full-width half-maximum (FWHM) kernel. The standardized ReHo values were then parcellated using the four different parcellation/segmentation schemes (Bianciardi et al., 2016; King et al., 2019; Schaefer et al., 2018; Tian et al., 2020), yielding 498 ReHo values for each fMRI segment. We used 400 cortical ROIs from the Schaefer atlas, 34 subcortical ROIs from the Tian atlas, 54 brainstem ROIs from the Bianciardi atlas, and 10 cerebellar ROIs from the MDTB atlas, for a total of 498 ROIs.

### 2.1.6 Data composition, units of analysis, and class imbalance

Three levels of organization are distinguished throughout. A participant is an individual meditator. A run is a single continuous fMRI acquisition, either a complete ACAM-J sequence or one control task. A segment is the portion of a run corresponding to a single labelled state, and it is the observation that enters the classifiers as one 498-dimensional ReHo feature vector. Because each participant contributes several runs and each run contributes several segments, observations are not independent of one another, and all model fitting and evaluation were organized so that participant identity, rather than segment count, defined the partitioning of the data (Section 2.4.1). Our dataset exhibits a marked imbalance: the case study participant contributed 194 segments across 27 ACAM-J runs and 16 runs each for the counting and memory tasks whereas the cohort study contributed only 161 entries. This corresponds to a 27:1.4 ratio of ACAM-J runs per participant (27 runs for the case-study participant vs. 1.4 runs per participant in the group). To address this, we employed both data-level and algorithm-level solutions.

#### *2.1.6.1 Data-level balancing with SMOTE*

We first applied the Synthetic Minority Over-sampling Technique (SMOTE), illustrated in **Fig. 1**, to the group study dataset (Chawla et al., 2002; Douzas & Bacao, 2019; He & Garcia, 2009). SMOTE generates new minority-class samples by interpolating existing instances, thus reducing class imbalance without simply replicating data points. SMOTE is a standard remedy for severe class imbalance and is used here as a training-time strategy only; the interpolated vectors it produces are not additional biological observations and are not interpreted as such at any point in this work.

Before applying SMOTE, certain ACAM-J and control tasks were severely underrepresented relative to the case study participant. For example, individual classes in the cohort often contained only a handful of samples compared to the much larger number contributed by the case study participant, contributing to the overall 27:1.4 imbalance. After applying SMOTE, training-fold classes were balanced at a 1:1 ratio, allowing the model to learn a richer decision boundary and mitigating overfitting to the majority class.

We first trained and selected models using only the group-study dataset, after balancing it with SMOTE. Specifically, (1) we split the group-study data into training and validation folds by participant, (2) within each fold we applied feature ranking, then RFE, then SMOTE, fitting each of these on the training portion only (Section 2.2), and (3) we trained candidate classifiers on this SMOTE-balanced training data and evaluated them on the unmodified validation data. This fold-wise procedure yields decision boundaries that are not dominated by majority classes while avoiding data leakage into validation. No synthetic or augmented observations were ever placed in a validation fold or in the held-out case-study set.

After cross-validation and model selection, we formed the final ensemble by combining the top-performing models (selected by validation performance) and refitting this ensemble on the full SMOTE-balanced group-study dataset. We then performed a single held-out evaluation by testing the final ensemble on the original, unmodified case-study dataset. This two-step strategy separates the balanced training distribution from the evaluation distribution, so that reported performance is not an artifact of resampling.

In practice, augmentation in the group-study training data was not limited to SMOTE alone. After SMOTE generated synthetic samples by interpolating between nearest neighbors in feature space, we further introduced controlled variability to avoid exact duplication and to better reflect plausible feature fluctuations. Concretely, we applied (i) light stochastic perturbations to numeric features using a noise_level parameter, (ii) blending of original and perturbed values controlled by 'retention_degree', and (iii) additional interpolation applied to a subset of numeric columns controlled by 'interpolation_proportion'. These steps, whose parameter values are listed in Supplementary Section S3, were applied only to training folds; the case-study test set remained completely unmodified.

## 2.2 Feature selection

We first selected a subset of ROIs from the 498 parcellated ROIs based on three key criteria: (1) prior results demonstrating significant trends across ACAM-J and clear differentiation from non-meditative control conditions (Yang et al., 2024a); (2) regions exhibiting reliable and stable ReHo values as determined through network intra-class correlation analyses (Ganesan et al., 2024); and (3) ROIs with established functional relevance to ACAM-J, particularly those involved in attentional regulation, sensory processing, and self-referential cognition (Yang et al., 2024b). This approach focused our analysis on regions with both empirical and theoretical significance to ACAM-J. This first step is literature-driven and does not use the labels of the present dataset. We then applied Recursive Feature Elimination (RFE) with a Random Forest model to identify the most informative neural features distinguishing ACAM-J from control tasks (Priyatno & Widiyaningtyas, 2024). RFE was chosen for its ability to consider feature interactions, and effectiveness with noisy and complex data (Breiman, 2001; Guyon et al., 2002). Random Forest was preferred for its resistance to overfitting, and ability to handle nonlinear relationships in small and variable datasets (Breiman, 2001; Cao & Shen, 2022; Zhao et al., 2016). Features with normalized importance scores below 0.5 were removed iteratively until only the most relevant features remained. The data-driven ranking step and all RFE model fits were computed on the training portion of each cross-validation fold. As detailed in Supplementary Section S2, the point at which elimination was stopped was chosen using performance on the same fold's validation data, so cross-validated metrics should be read as internal model-selection estimates rather than as fully independent validation. The held-out case-study data were not used at any stage of feature ranking, elimination, class balancing, model fitting, or model selection.

### *2.3 Machine learning module*

#### *2.3.1 ML Classifier families*

We applied several established ML models to classify ACAM-J, chosen for their suitability to small-sample, high-dimensional neuroimaging data. None of these algorithms is novel; they are used here as analytical tools for decoding. Logistic regression models the probability of categorical outcomes through a logistic function and yields interpretable coefficients that indicate the strength and direction of the relationship between each feature and the predicted outcome (Freedman, 2009). Decision trees partition the feature space through hierarchical decision rules, offering an intuitive view of which regions drive a classification (Quinlan, 1986). Random forest, an ensemble of decision trees trained on bootstrapped datasets, improves stability by averaging across trees and provides impurity-based importance metrics useful for identifying candidate neurobiological correlates (Breiman, 2001). Support vector machines with linear kernels separate classes using a hyperplane optimized for high-dimensional data, require minimal tuning, and remain straightforward to interpret (Cortes & Vapnik, 1995). k-nearest neighbors (kNN) classifies samples by the majority class among nearest neighbors in feature space and makes no parametric assumptions about the decision surface (Cover & Hart, 1967). Artificial neural networks, here multi-layer perceptrons, model nonlinear relationships between input and output layers and can capture distributed patterns that linear models miss (Rumelhart et al., 1986).

Several classifiers were deliberately excluded. Gradient boosting machines and related methods such as XGBoost, AdaBoost, and LightGBM (Chen & Guestrin, 2016; Friedman, 2001) were not used because an ensemble strategy is already applied at the model-combination stage. Naive Bayes and non-linear SVM kernels were not used given the characteristics of our dataset and our focus on methodological simplicity (McCallum & Nigam, 1998). Deep learning was omitted because of the small sample size and the associated overfitting risk (Krizhevsky et al., 2017). Clustering, dimensionality reduction, and reinforcement learning were not aligned with the aims of this study (Sutton & Barto, 2018).

### 2.3.2 Model training with stratified cross-validation

We used subject-wise stratified 5-fold cross-validation to train the SMOTE-balanced dataset (Kohavi, 1995). Folds were formed so that all segments and runs contributed by a given participant fall entirely within either the training or the validation portion of a fold, and never both, while class proportions were preserved in each fold (Chawla et al., 2002; Kubat et al., 1998). Partitioning by participant rather than by segment means that validation performance reflects prediction for participants whose data the model has not seen, rather than for further segments drawn from participants already in the training set. Stratification also ensures that class distributions remain representative across folds, which stabilizes performance estimates under imbalance (Buda et al., 2018; He & Garcia, 2009).

### 2.3.3 Handling overfitting

To mitigate and assess overfitting risks, we introduced diversity through SMOTE and leveraged stratification to keep class ratios consistent across folds. We also used a broad set of evaluation metrics, reporting chance-adjusted agreement (Cohen's kappa) and discrimination (AUC) alongside accuracy, and testing observed performance against a permutation-based null distribution (Section 2.4.2) rather than relying on any single index. Emphasis was placed on held-out performance to assess the models' applicability to unseen data. Lastly, we prepared Early Stopping (Alberti et al., 2017) to halt training should validation performance degrade after a peak; however, no such decline occurred, so this safeguard was never triggered.

#### 2.3.4 Ensemble learning

To further improve predictive performance and reduce variance through the aggregation of multiple learners, we implemented ensemble learning by combining the predictions of multiple ML models (Dietterich, 2000; Polikar, 2006). After training all candidate classifiers, we selected the top three models by validation Cohen's κ, with ties broken by higher AUC (Supplementary Section S4), and averaged their predicted probabilities to generate final ensemble outputs at a decision threshold of 0.5. Probability averaging was chosen for its simplicity, transparency, and stability (Polikar, 2006). While other ensemble strategies like weighted averaging or majority voting could be used, probability averaging provides a balanced and interpretable solution by integrating model confidence into the final prediction without requiring additional hyperparameters or weighting schemes. Ensemble outputs were then evaluated with the metrics described in Section 2.4.2.

### 2.4 Model testing and evaluation

#### *2.4.1 Testing strategy and parameter setting*

Models were evaluated at two distinct levels, and we distinguish these throughout. All models were first trained and validated exclusively on the group-study dataset using subject-wise stratified 5-fold cross-validation, with SMOTE applied to training folds only. These cross-validated metrics describe sample-level predictive performance and, because folds are partitioned by participant, provide a first estimate of performance for unseen participants within the group cohort. After model selection, the final ensemble was tested once on the unmodified case-study dataset. This second evaluation is an intensive single-case, cross-participant generalization test: it asks whether a model fitted on the group cohort transfers to densely sampled data from a different advanced practitioner, on the natural, unbalanced distribution of ACAM-J states and control conditions. It does not, on its own, establish population-level generalization, which would require a larger and more demographically diverse independent sample. Hyperparameters were fixed a priori at the values listed in Supplementary Table S1 and were held constant across folds, contrasts, and the final case-study evaluation. Following model selection, ensemble learning was applied by averaging predicted probabilities from the three best-performing models.

#### *2.4.2 Evaluation metrics*

We assessed model performance using a complementary set of classification metrics to capture different aspects of predictive accuracy and reliability. Overall effectiveness was quantified using accuracy, while Cohen’s kappa provided a chance-adjusted measure of agreement between predicted and true labels (Cohen, 1960). Discriminative ability across decision thresholds was assessed using the area under the ROC curve (AUC). Precision and recall (sensitivity) captured the model's reliability in identifying positive instances, and the F1 score summarized their balance under class imbalance. Specificity was included to evaluate the correct identification of negative cases, thereby accounting for false-positive risk. Statistical significance was assessed with a subject-wise permutation test (1,000 iterations) in which class labels were randomly permuted within each participant, the full pipeline was re-run, and the observed metric was compared against the resulting right-tailed null distribution (Supplementary Section S2). The reported p-values therefore indicate how unlikely the observed level of performance would be under a null in which ReHo patterns carry no information about condition labels. They do not by themselves establish that a classifier is reliable at the individual level or that it will generalize to new populations; accuracy, chance-adjusted agreement, discrimination, and the permutation test are reported together and interpreted jointly.

### *2.5 Pairwise feature significance analysis of ACAM-J and control tasks*

To characterize how neural features differ with ACAM-J depth, we analyzed transitions between successive ACAM-J (e.g., ACAM-J1 to ACAM-J2, ACAM-J2 to ACAM-J3), as well as comparisons between ACAM-J and the non-meditative control tasks. These analyses capture both the gradual progression of concentrative absorption and the contrast between ACAM-J and baseline cognitive processes. We also examined a targeted comparison between ACAM-J1 and ACAM-J6 to capture the difference between early and late ACAM-J. This widely separated contrast provides a stringent test of the model's capacity to discriminate early, content-rich states from advanced, highly abstract states, and allows adjacent and distant state pairs to be compared directly.

Feature significance was assessed using model-appropriate methods. For tree-based models (e.g., random forests, gradient boosted trees, decision trees), we used impurity-based feature importance to quantify each variable's contribution to split quality (Breiman, 2001; Pedregosa et al., 2011). For linear models such as logistic regression and linear-kernel SVMs, feature coefficients were extracted to indicate the direction and magnitude of each ROI's influence on classification (Pedregosa et al., 2011). For models without built-in feature impact metrics, such as *k*-Nearest Neighbors, Neural Networks, or SVC with non-linear kernels, we applied "permutation importance". This model-agnostic method evaluates a feature's importance by noting the change in model performance when the values of that feature are randomly shuffled (Altmann et al., 2010; Fisher et al., 2019). All importance values were normalized to a 0-1 range to standardize interpretability across model types and to enable direct comparison of the most discriminative features. These values quantify each ROI's contribution to classifier decisions and are not statistical tests of ReHo differences between conditions; the two are displayed separately in the figures. For visualization, only ROIs meeting the permutation-based significance criterion consistently across all cross-validation folds were retained in the brain maps (Figs. 4 and 5). In addition, we quantified how often each ROI was necessary for classification using leave-one-region-out analyses, in which each ROI's ReHo feature was removed in turn and the number of classifiers whose accuracy dropped was recorded across the ACAM-J vs control, counting, and memory comparisons. The leave-one-region-out analysis was performed across the ACAM-J vs. control, ACAM-J vs. counting, and ACAM-J vs. memory contrasts using the six classifier families described in Section 2.3.1. The resulting counts are displayed as ROI overlap maps (Section 3.3).

### *2.6 Additional control analysis*

Given that phenomenological experiences were associated with brain activity during ACAM-J (Ganesan et al., 2024; Potash et al., 2024), we performed additional analyses to account for individual differences in phenomenology. We used linear regression to regress out phenomenological measures from ReHo values. For each ROI, ReHo was treated as the dependent variable, and phenomenological variables were entered as predictors. Predictors included stability of attention and width of attention for both ACAM-J and control tasks, as well as intensity of ACAM-J for the ACAM-J task; early and late phenomenology items were not included. Residualization parameters were fitted within training folds and applied to validation and case-study data (Supplementary Section S2). Residual ReHo values were used as input to the same machine learning pipeline described above. This analysis asks whether classification performance changes once variance associated with reported phenomenological dimensions is accounted for; it does not isolate a phenomenology-free component of ACAM-J neural activity.

## 3 Results

### 3.1 ACAM-J vs control conditions

We first evaluated the broad distinction between ACAM-J and non-ACAM-J. ReHo differences between these categories reveal distinct functional signatures, with the most discriminative regions spanning the default mode network (medial prefrontal and posterior parietal cortices), anterior insula, anterior cingulate cortex, and the locus coeruleus (**Fig. 2**). Using machine-learning classifiers, the model achieved an accuracy of 73.19% and a Cohen's kappa (κ) of 0.2604 for ACAM-J vs Counting, and 63.94% accuracy and κ of 0.1908 for ACAM-J vs Memory.

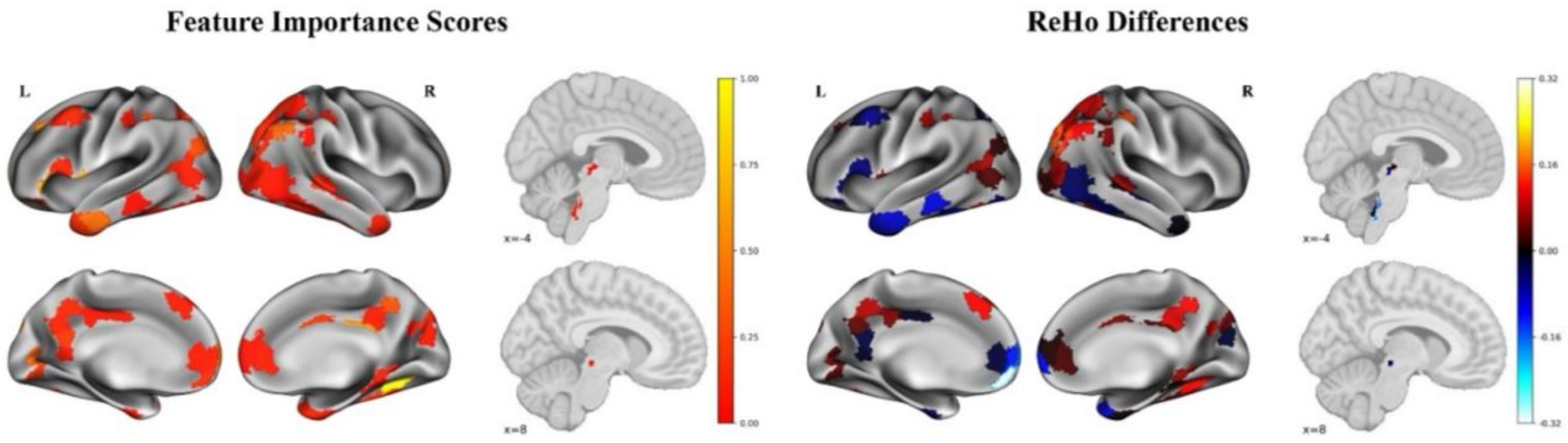

***Fig. 2: Brain-plots for ACAM-J vs control (counting and memory combined) conditions.*** *Left panels display normalized feature-importance scores for the cortical and subcortical ROIs identified via the pairwise feature significance analysis, indicating each region's contribution to classifier decisions rather than a statistical test of condition differences. Right panels show whole-brain surface maps of the average ReHo difference between all ACAM-J and combined control (counting + memory) conditions, and the color bar indicates ReHo differences, with warm colors for higher ReHo during ACAM-J and cool colors the opposite.*

Next, we examined the contrast between ACAM-J and the counting task. As shown in Fig. 3, ReHo differences reveal distinct spatial patterns across the six ACAM-J stages relative to counting. Discriminative regions include the locus coeruleus, as well as distributed clusters across visual, control, default mode, and salience networks. Beginning with ACAM-J4, regions within the dorsal attention network become increasingly prominent, suggesting a shift toward more stable and sustained attentional engagement during mid-to-late ACAM-J. Detailed metrics for each ACAM-J stage, including accuracy, κ, AUC, precision, recall, F1 score, sensitivity, specificity, and *p*-values, are summarized in Table 1.

***Table 1****: Classification metrics for ACAM-J vs. Counting Task. The accuracy (Acc), Cohen's kappa (CK), Area under the ROC curve (AUC), Precision, Recall/Sensitivity, F1 (harmonic mean of precision and recall), Specificity (TNR) and p-values are derived from the subject-wise permutation test described in Section 2.4.2.*

| **Metric** | **ACAM-J vs Counting** | **ACAM-J1 vs Counting** | **ACAM-J2 vs Counting** | **ACAM-J3 vs Counting** | **ACAM-J4 vs Counting** | **ACAM-J5 vs Counting** | **ACAM-J6 vs Counting** |
|---|---|---|---|---|---|---|---|
| **Acc** | 0.7319 | 0.7044 | 0.6524 | 0.7150 | 0.5363 | 0.7086 | 0.6618 |
| **CK** | 0.2604 | 0.2882 | 0.2609 | 0.2739 | 0.0875 | 0.3306 | 0.3278 |
| **AUC** | 0.7892 | 0.7536 | 0.6903 | 0.7693 | 0.5313 | 0.7616 | 0.7027 |
| **Precision** | 0.7415 | 0.7213 | 0.6689 | 0.7346 | 0.5447 | 0.7298 | 0.6821 |
| **Recall / Sens.** | 0.7368 | 0.7142 | 0.6613 | 0.7284 | 0.5412 | 0.7244 | 0.6765 |
| **F1** | 0.7396 | 0.7177 | 0.6651 | 0.7315 | 0.5429 | 0.7271 | 0.6793 |
| **Specificity** | 0.6270 | 0.6946 | 0.6435 | 0.7016 | 0.5314 | 0.6928 | 0.6471 |
| **p-Value** | <0.001 | <0.001 | <0.001 | <0.001 | 0.031 | <0.001 | <0.001 |

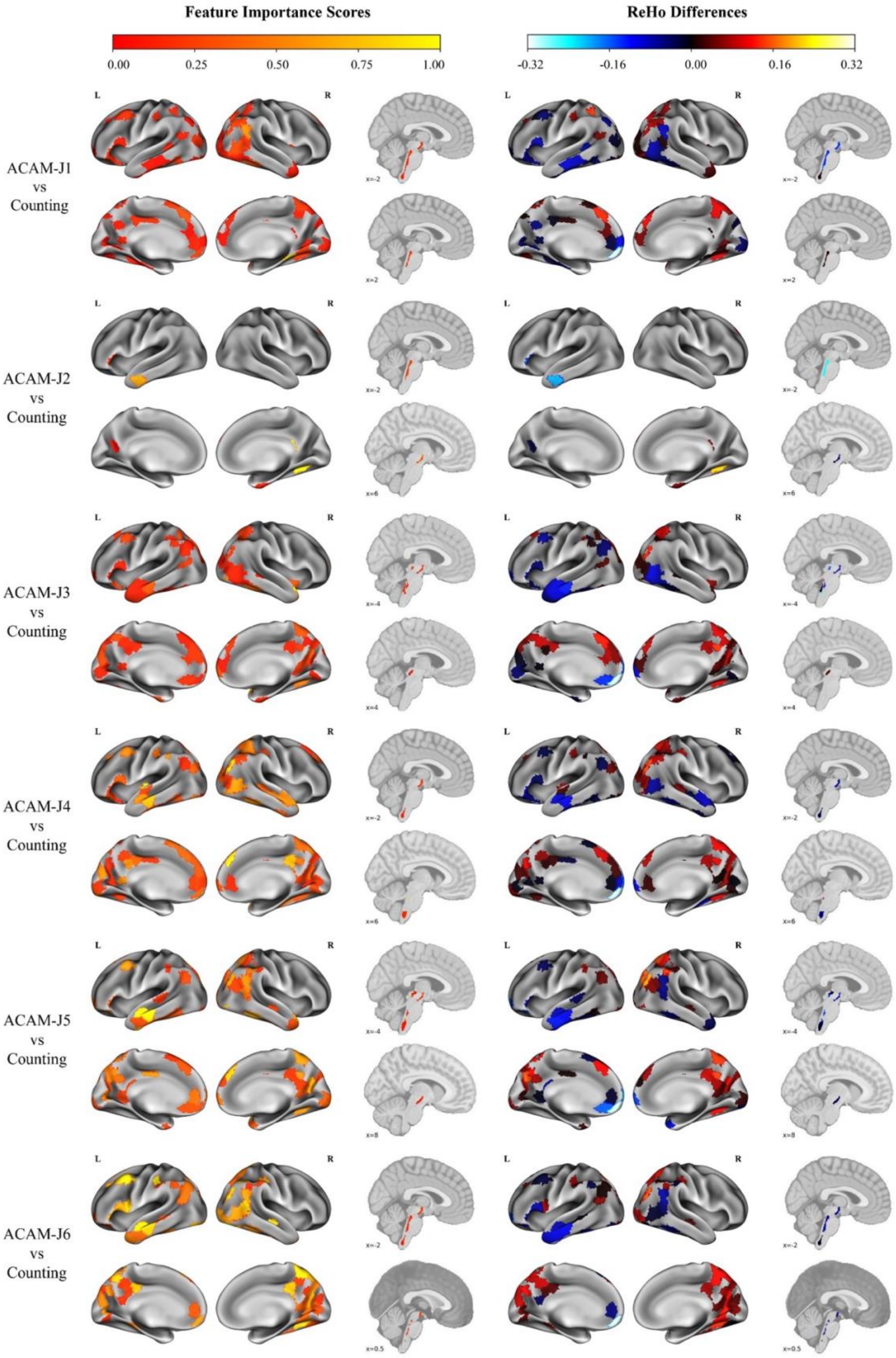

***Fig. 3: Brain-plots for ACAM-J vs. counting.*** *Left panels display normalized feature-importance scores for the cortical and subcortical ROIs identified via the pairwise feature significance analysis. Right panels show six sagittal slices of the average ReHo difference for each comparison, with the color bar indicating ReHo differences, with warm colors indicating higher local synchronization during ACAM-J and cool colors the opposite.*

When distinguishing ACAM-J from the memory task, discriminability was more modest than for counting (Table 2). The brain plots and detailed performance metrics for each ACAM-J stage relative to memory are provided in Fig. 4 and Table 2 respectively. ACAM-J exhibits increased ReHo within default mode and control networks, as well as salience network regions and the locus coeruleus. These regions become especially prominent in ACAM-J5 and ACAM-J6. Unexpected neural signatures were observed particularly for ACAM-J2 versus counting and memory, highlighting potential neurophenomenological distinctions that warrant further investigation. Discriminability was also not uniform across states: the ACAM-J4 contrasts were the weakest against both control tasks, approaching chance for ACAM-J4 vs Memory (accuracy 0.4937, AUC 0.5011, $\kappa = 0.1102$) and for ACAM-J4 vs Counting (accuracy 0.5363, AUC 0.5313, $\kappa = 0.0875$). This pattern is considered further in Section 4.2.

***Table 2****: Classification metrics for ACAM-J vs. Memory. The accuracy (Acc), Cohen's kappa (CK), Area under the ROC curve (AUC), Precision, Recall/Sensitivity, F1 (harmonic mean of precision and recall), Specificity (TNR) and p-values are derived from the subject-wise permutation test described in Section 2.4.2.*

| **Metric** | **ACAM-J vs Memory** | **ACAM-J1 vs Memory** | **ACAM-J2 vs Memory** | **ACAM-J3 vs Memory** | **ACAM-J4 vs Memory** | **ACAM-J5 vs Memory** | **ACAM-J6 vs Memory** |
|---|---|---|---|---|---|---|---|
| **Acc** | 0.6394 | 0.6475 | 0.5824 | 0.7034 | 0.4937 | 0.6667 | 0.7430 |
| **CK** | 0.1908 | 0.1928 | 0.1329 | 0.2406 | 0.1102 | 0.1521 | 0.3164 |
| **AUC** | 0.6694 | 0.6743 | 0.5959 | 0.7460 | 0.5011 | 0.6993 | 0.7937 |
| **Precision** | 0.6587 | 0.6629 | 0.5946 | 0.7215 | 0.5047 | 0.6799 | 0.7653 |
| **Recall / Sens.** | 0.6703 | 0.6749 | 0.6045 | 0.7353 | 0.5137 | 0.6905 | 0.7818 |
| **F1** | 0.6616 | 0.6689 | 0.5995 | 0.7283 | 0.5091 | 0.6852 | 0.7734 |
| **Specificity** | 0.6113 | 0.6199 | 0.5603 | 0.6715 | 0.4737 | 0.6429 | 0.7042 |
| ***p*-Value** | <0.001 | <0.001 | 0.027 | <0.001 | 0.029 | <0.001 | <0.001 |

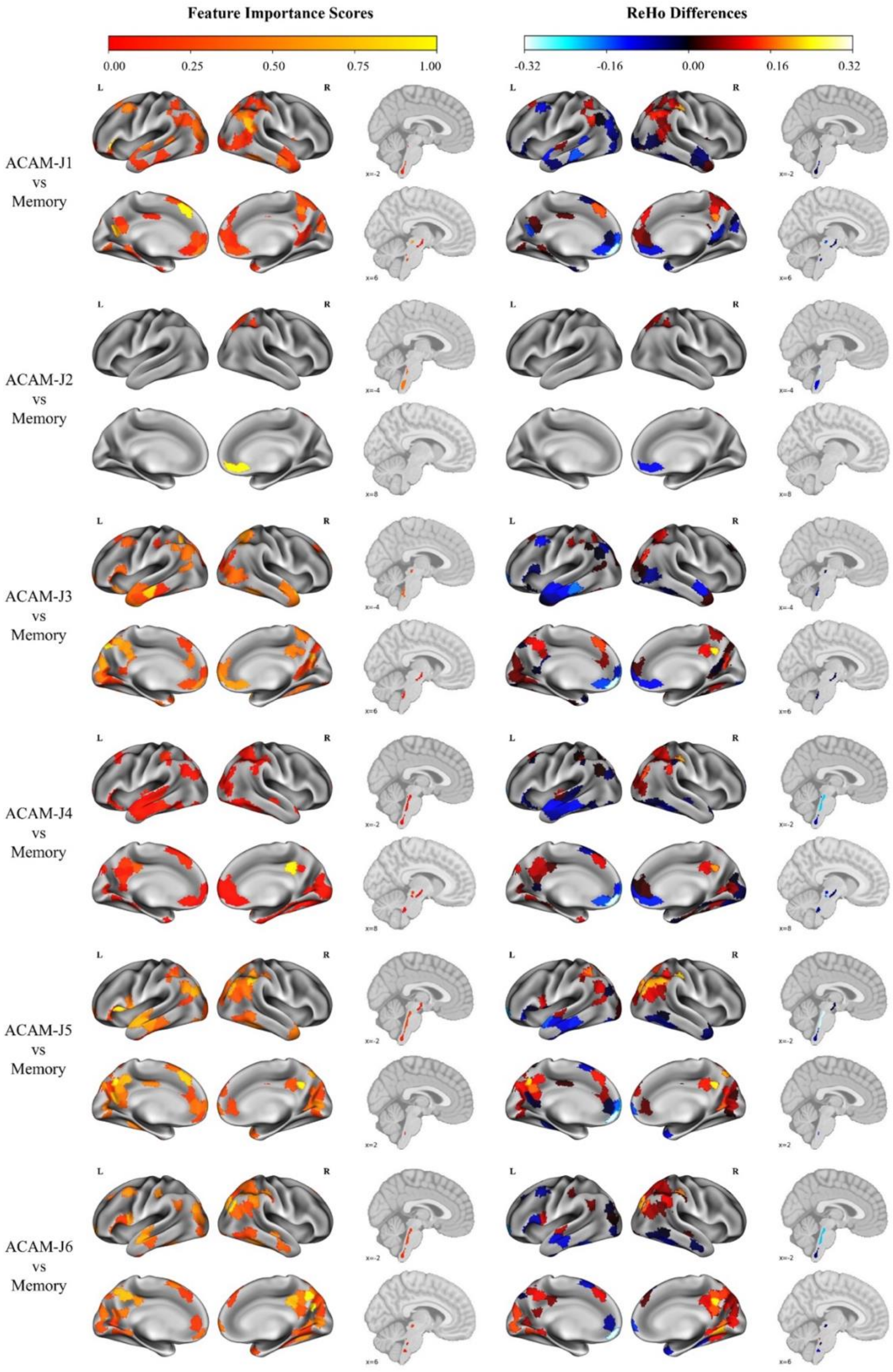

***Fig. 4: Brain-plots for ACAM-J vs. Memory task.*** *Left panels display normalized feature-importance scores for the cortical and subcortical ROIs identified via the pairwise feature significance analysis. Right panels show sagittal slices of the average ReHo difference between each comparison, with the color bar indicating ReHo differences, with warm colors marking increased synchronization in ACAM-J. Only ROIs selected by permutation-based significance across all folds are displayed, emphasizing consistent discriminators within default-mode, control, salience networks, and the locus coeruleus (see Table 2 for evaluation metrics).*

### 3.2 Pairwise ACAM-J classification

Pairwise classification between adjacent ACAM-J stages provides a finer-grained view of neurophenomenological transitions (Table 3; Fig. 5). Across all ACAM-J pairs, the model achieved an overall accuracy of 66.48% with a Cohen's κ of 0.2779, indicating modest but meaningful discriminability among closely related states. Performance varied systematically with the separation between states: the widely separated ACAM-J1 vs ACAM-J6 contrast was the most discriminable (74.33% accuracy, κ = 0.5158, AUC = 0.8433), followed by ACAM-J4 vs ACAM-J5 (71.89%, κ = 0.4613), whereas adjacent early transitions such as ACAM-J2 vs ACAM-J3 remained close to the lower end of the range (61.68%, κ = 0.1461). ReHo contrasts reveal progressive shifts in ReHo associated with ACAM-J depth. Key discriminative regions include hubs of the default mode network, particularly medial prefrontal areas, along with the cingulate, control network nodes, salience network regions, and locus coeruleus. These regions become especially prominent from ACAM-J4 onward.

***Table 3**: Classification metrics for successive ACAM-J pairs and ACAM-J1 vs ACAM-J6. The accuracy (Acc), Cohen's kappa (CK), Area under the ROC curve (AUC), Precision, Recall/Sensitivity, F1 (harmonic mean of precision and recall), Specificity (TNR) and p-values are derived from the subject-wise permutation test described in Section 2.4.2.*

| **Metric** | **ACAM-J1 vs ACAM-J2** | **ACAM-J2 vs ACAM-J3** | **ACAM-J3 vs ACAM-J4** | **ACAM-J4 vs ACAM-J5** | **ACAM-J5 vs ACAM-J6** | **ACAM-J1 vs ACAM-J6** |
|---|---|---|---|---|---|---|
| **Acc** | 0.6319 | 0.6168 | 0.6559 | 0.7189 | 0.6222 | 0.7433 |
| **CK** | 0.2270 | 0.1461 | 0.1470 | 0.4613 | 0.1704 | 0.5158 |
| **AUC** | 0.6918 | 0.6770 | 0.7194 | 0.8049 | 0.6596 | 0.8433 |
| **Precision** | 0.6524 | 0.6391 | 0.6967 | 0.7838 | 0.6482 | 0.8267 |
| **Recall / Sens.** | 0.6568 | 0.6457 | 0.7079 | 0.7996 | 0.6571 | 0.8394 |
| **F1** | 0.6546 | 0.6424 | 0.7022 | 0.7916 | 0.6526 | 0.8330 |
| **Specificity** | 0.6306 | 0.6227 | 0.6477 | 0.7174 | 0.6065 | 0.7666 |
| ***p*-Value** | <0.001 | <0.001 | <0.001 | 0.011 | <0.001 | 0.009 |

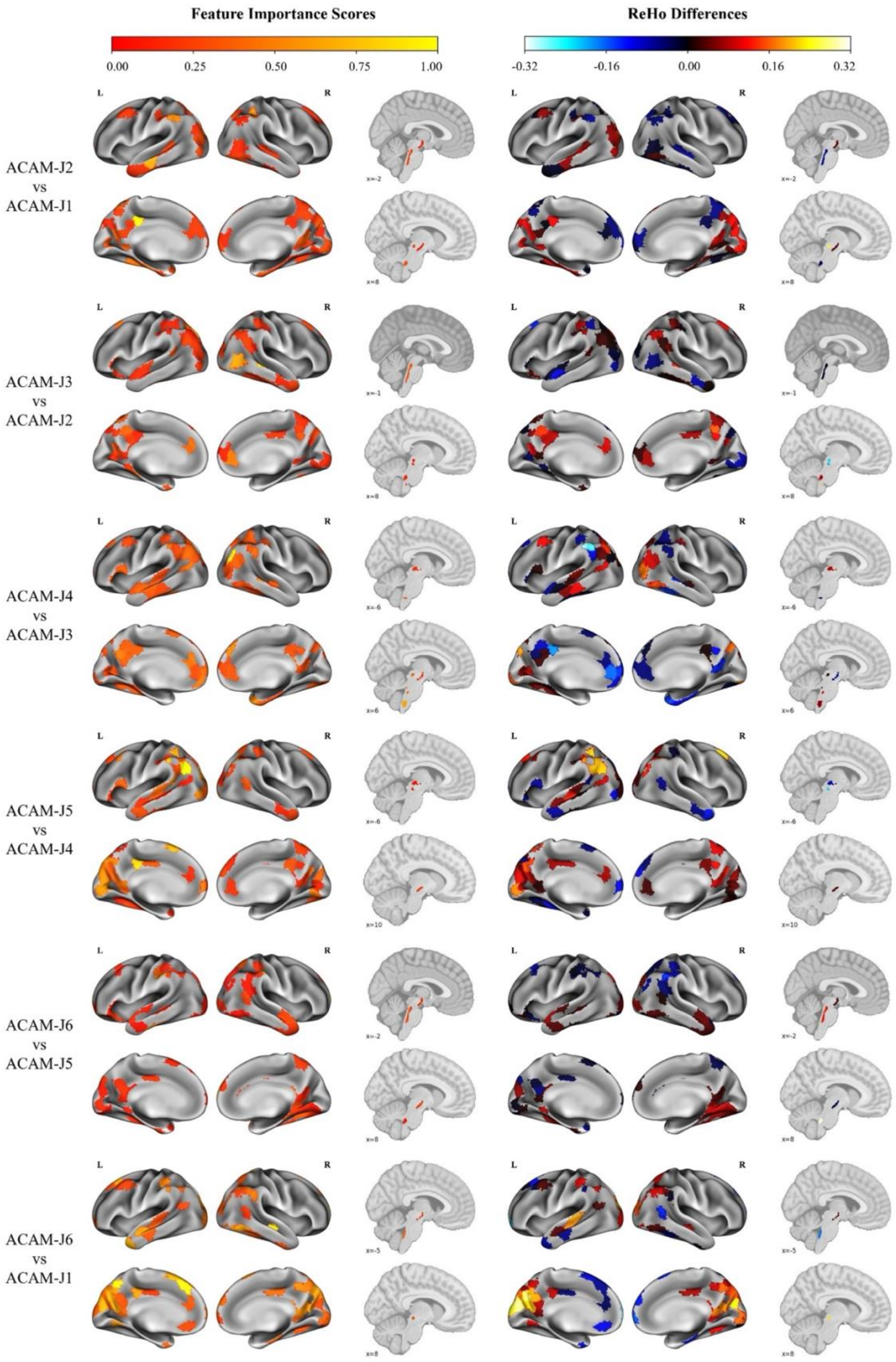

***Fig. 5: Brain-plots for pairwise ACAM-J transitions.*** *Left panels display normalized feature-importance scores for the cortical and subcortical ROIs identified via the pairwise feature significance analysis. Right panels show sagittal slices of the average ReHo difference between each comparison, with the color bar indicating ReHo differences, with warm colors marking increased synchronization in ACAM-J. Only ROIs selected by permutation-based significance across all folds are displayed, emphasizing consistent discriminators within default-mode, control, salience networks, and the locus coeruleus (see Table 3 for evaluation metrics).*

### 3.3 ROI overlap maps

To further delineate which brain regions were most consequential for differentiating ACAM-J from non-meditative control conditions and cognitive tasks, ROI overlap analysis was conducted based on leave-one-region-out analyses (Section 2.5). Fig. 6 presents these overlap maps, identifying regions whose removal reduced classification accuracy across the largest number of classifiers, which points to the prefrontal cortex, anterior cingulate cortex, insula, and parietal lobes, areas associated with attention, emotional processing, interoceptive awareness, and spatial cognition, respectively. The prominence of Prefrontal Cortex, involved in attention and executive function (Kerr et al., 2013), Anterior Cingulate Cortex, associated with self-regulation and emotional processing (Sparby & Sacchet, 2022), Insula, linked to interoceptive awareness and salience tagging (Giommi et al., 2023), Parietal Lobes, related to spatial awareness and attentional orienting, and Locus Coeruleus, implicated in noradrenergic arousal modulation (Abellaneda-Pérez et al., 2024), suggests that these hubs jointly underpin the discriminative neural patterns observed across ACAM-J.

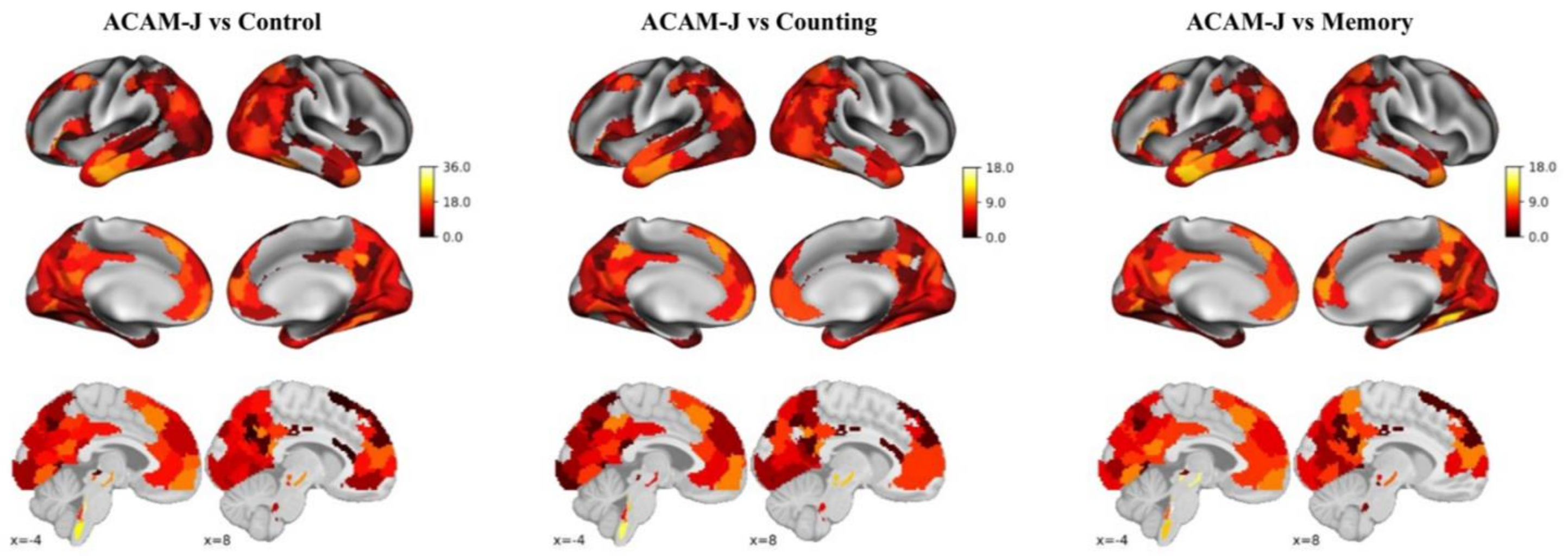

***Fig. 6: ROI Overlap Maps with cortical (above) and subcortical (below) plots.*** *Color intensity indicates, for each ROI, the number of classifiers whose accuracy dropped when that region's ReHo feature was removed in leave-one-region-out analyses across ACAM-J vs. control, counting, and memory comparisons. Brighter colors (prefrontal, cingulate, insula, parietal) identify regions most critical to classification performance.*

### 3.4 Residual Analysis

Classification on the residualized data changed the overall average accuracy from 65.87% to 66.82%, a difference of 0.95 percentage points. The overall average Cohen's κ before residualization was 0.2443. This small shift indicates that classification did not depend on variance shared with the reported phenomenological dimensions, and that accounting for that variance left discriminability essentially intact or marginally improved; it does not establish that the residual signal reflects ACAM-J-specific neural activity alone (Ganesan et al., 2024; Potash et al., 2024). The changes in accuracy and Cohen's Kappa scores before and after residual analysis are shown in **Fig. 7**.

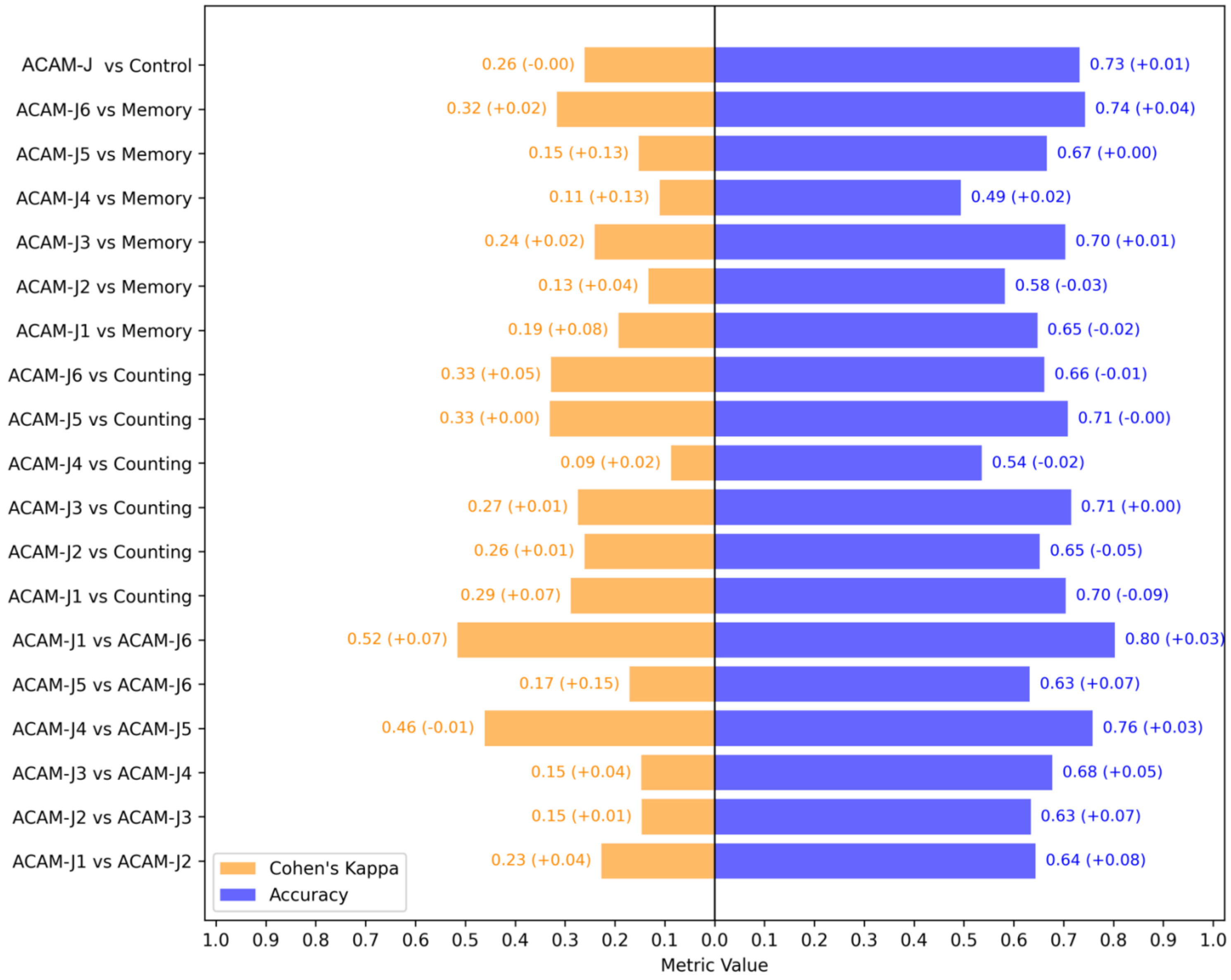

***Fig. 7: Summary of Accuracy and Cohen's Kappa Scores.*** *Clustered bars display overall accuracy and Cohen's κ for each binary comparison, with bracketed annotations indicating the change in Cohen's Kappa and Accuracy resulting from removing phenomenological variance via residual analysis.*

### 3.5 Summary of the results

Collectively, the machine learning classification analysis across the 19 binary comparisons summarized in Fig. 7 yielded an overall average accuracy of 66.82% after residualization (65.87% before residualization; Section 3.4) and an overall average Cohen's Kappa of 0.2443 before residualization, with all comparisons exceeding their permutation null at $p < 0.05$. The integrated findings summarized visually in Fig. 7 indicate that ReHo-derived neural patterns carry information that discriminates between ACAM-J and non-meditative conditions, and, with lower but above-null performance, between ACAM-J states themselves.

## 4 Discussion

In this study, we applied a machine learning pipeline, informed by first-person phenomenological reports, to distinguish ACAM-J from cognitively active control conditions and to discriminate among progressive ACAM-J states using 7T fMRI-derived ReHo. To our knowledge, this is the first demonstration of predictive models that decode ACAM-J states from MRI data; the claim is limited to ACAM-J and to MRI and is not a claim of priority for machine learning in meditation research more broadly. Beyond establishing the feasibility of ACAM-J decoding, the analytical approach is applicable to other rare or difficult-to-sample cognitive states. The candidate neural signatures identified here may inform future translational work, including neurofeedback and individualized monitoring of meditative or therapeutic training, although such applications are not demonstrated by the present design.

### 4.1 Tailored machine learning pipeline for rare cognitive states

Rather than treating rarity as a methodological obstacle, our approach treats it as a defining property of the cognitive phenomena under investigation. ACAM-J is not only uncommon in the general population, but is also characterized by subtle, internally stabilized experiential dynamics that unfold over training and practice. Capturing such states therefore requires analytical frameworks optimized for precision, interpretability, and robustness under constrained sampling conditions (He & Garcia, 2009; T R M et al., 2023).

To this end, we assembled a pipeline from established components (He & Garcia, 2009; T R M et al., 2023), including the Synthetic Minority Oversampling Technique (SMOTE), subject-wise stratified cross-validation, RFE, and an ensemble-based approach. None of these components is new; their contribution here is to make decoding of a rare state tractable while limiting optimistic bias. SMOTE was employed as a training-time strategy for stabilizing decision boundaries in underrepresented absorption states rather than as a source of additional observations (He & Garcia, 2009; T R M et al., 2023). Stratification maintained the class distributions of ACAM-J across folds, while participant-wise partitioning ensured that validation folds contained only participants absent from training.

Interpretability was treated as a core design constraint rather than a post hoc add-on. RFE enabled identification of the brain regions contributing most strongly to classification performance without imposing strict assumptions about feature dominance or hierarchical ordering.

Ensemble learning reduced sensitivity to the idiosyncrasies of individual models by aggregating predictions across classifiers. Early stopping was available as a safeguard but was never triggered, since validation performance did not decline after peaking (Section 2.3.3). The consistent appearance of k-Nearest Neighbors, Logistic Regression, and Neural Networks among the top-ranked models (Supplementary Section S4) reflects the adaptability of these models to the complex, high-dimensional nature of neuroimaging data. Taken together, this pipeline illustrates how carefully constrained machine-learning strategies can reveal above-chance neural distinctions even in small samples and rare cognitive states such as ACAM-J.

### 4.2 A neural model of advanced concentrative absorption

Our findings are consistent with a network-level account of profound states of concentrative absorption characterized by a dynamic interaction between large-scale brain networks (Potash et al., 2025; Treves et al., 2025). The discriminative features identified across ACAM-J included the brainstem locus coeruleus (LC), salience, control (e.g., prefrontal cortex and cingulate), and default mode (e.g., medial prefrontal, precuneus, and posterior cingulate cortex) networks.

A key element of this account is the recalibration of arousal and attention coupling. Increased LC contribution at later ACAM-J aligns with its established role in arousal regulation and with ultra-high-field MRI evidence characterizing structural connectivity of brainstem autonomic nuclei in living humans (K. Singh et al., 2022). Hence, LC contribution may reflect a refined autonomic state of stable wakefulness with minimal arousal, consistent with accounts of "effortless alertness" in advanced concentrative absorption (Yang et al., 2025). In parallel, the salience network becomes increasingly prominent, suggesting that practitioners increasingly rely on this system to maintain precise contact with the meditative object while minimizing intrusion of mental proliferation (Giommi et al., 2023; Kerr et al., 2013). The contribution of salience-related regions is not indicative of effortful control, but rather of a finely tuned detection-and-stabilization process. This interpretation aligns with neurophenomenological evidence that deep absorption depends not on effortful suppression, but on the stable maintenance of an attentional attractor state (Yang et al., 2024a; Yang et al., 2025).

In contrast, regions of the anterior DMN become less central in distinguishing deeper ACAM-J. This attenuation is consistent with reduced self-referential processing, decreased language-related coupling, and diminished spontaneous mentation documented in long-term practitioners (Sorella et al., 2025; Tripathi et al., 2024). Notably, this reduction is already evident by ACAM-J4, which phenomenologically marks the culmination of form-based absorption (Yang et al., 2024b). At this stage, self-referential activity appears to be largely minimized, suggesting that the decoupling of narrative self-processing precedes the transition to formless absorption rather than emerging only at later, more abstract ACAM-J. Control networks, including dorsolateral prefrontal and cingulate cortices, remain integral for sustaining attention (Yordanova et al., 2021), an important characteristic of ACAM-J. Together, these network shifts are compatible with a model in which concentrative absorption emerges from heightened alertness, strengthened object-based salience, sustained attentional stability, and reduced self-generated cognition.

The pairwise results bear on this account in a specific way. Discrimination was clearest for the most widely separated states, ACAM-J1 vs ACAM-J6, and for ACAM-J4 vs ACAM-J5, the transition from form-based to formless absorption, whereas several adjacent early transitions were harder to separate (Table 3). This pattern indicates that distributed ReHo patterns carry information about differences between advanced meditative states, and that the amount of that information is greater for states that practitioners describe as more distinct. Classification performance alone does not demonstrate a continuous neural trajectory through the ACAM-J sequence, and we did not model such a trajectory here; establishing one would require analyses designed for ordinal or dynamic structure. The ACAM-J4 contrasts against both control tasks were close to chance (Tables 1 and 2), so the current data do not support the claim that every individual ACAM-J state carries a separable ReHo signature; this uneven profile is a candidate observation for follow-up rather than an established feature of the sequence.

Incorporating self-reported phenomenological data into the classification pipeline changed overall average accuracy by 0.95 percentage points, from 65.87% to 66.82%. Two readings are consistent with this result: that classification relies largely on variance not shared with the measured phenomenological dimensions, and that residualization removes a modest amount of noise associated with individual differences in reported experience. We do not interpret the residualized signal as a phenomenology-free index of ACAM-J. What the analysis does support is the practical value of collecting structured subjective reports alongside neural measures in studies of complex cognitive states (Galante et al., 2023; Ganesan et al., 2024; Sparby & Sacchet, 2022).

### 4.3 Potential applications

The data-driven framework introduced here suggests several directions for future translational work, none of which is demonstrated by the present study. Neural signatures identified for ACAM-J may inform targeted non-invasive neuromodulation paradigms such as neurofeedback protocols (Abellaneda-Pérez et al., 2024). For example, real-time feedback of activity in posterior and prefrontal regions could in principle help individuals facing attentional lapses to progressively attain advanced concentration, and could support practitioners in stabilizing concentration. Non-invasive neuromodulation (e.g., TMS, tDCS) could be deployed to modulate arousal, salience, or control systems identified by the model, enabling mechanistic tests of how these networks contribute to sustained absorption (Abellaneda-Pérez et al., 2024). When combined with real-time classification, such interventions would lend themselves to closed-loop designs that synchronize neural data with first-person reports, allowing moment-to-moment tracking of state transitions (Galante et al., 2023; Lutz et al., 2007; Potash et al., 2024; Van Lutterveld et al., 2017). More broadly, embedding ACAM-J-derived measures into longitudinal clinical and contemplative trials could allow quantification of dose and response relationships, individual learning trajectories, and inter-individual variability in absorption capacity (Galante et al., 2023). Each of these applications would first require validation in larger and more diverse samples, and at the temporal resolution that real-time use demands.

### 4.4 Challenges and Future Directions

A key limitation of this study is the relatively small number of participants capable of entering and maintaining ACAM-J. Later ACAM-J, in particular, are rare and challenging to sustain, resulting in fewer usable data samples. Because each participant contributes many runs and segments, the effective number of independent observations is far smaller than the number of feature vectors entering the models, and the reported metrics should be read as sample-level predictive performance under participant-wise partitioning rather than as population-level generalization. The final evaluation on the intensively sampled case participant is an informative cross-participant test, but it is a test on one additional individual and cannot establish that the models transfer to advanced practitioners in general. The group cohort was also predominantly male (19 of 20 participants) and self-selected from established meditation communities, which further constrains the populations to which these results extend. Although data-balancing strategies such as SMOTE helped mitigate class imbalance (Chawla et al., 2002; He & Garcia, 2009), synthetic augmentation cannot fully recapitulate the nuanced, heterogeneous neural signatures associated with advanced meditative absorption. As a result, models trained on SMOTE-enhanced datasets remain at risk of overfitting and may not always generalize to new participants or sessions. A further methodological caveat is that the stopping point of recursive feature elimination was selected using validation-fold performance within the same fold (Section 2.2), so cross-validated metrics are best treated as model-selection estimates; a fully nested selection loop would provide stricter estimates in future work.

These constraints also informed our decision to prioritize classical, interpretable machine learning methods over deep learning approaches, which typically require substantially larger datasets and risk collapsing under small-N conditions (Krizhevsky et al., 2017). While this choice improves reproducibility and transparency, it also caps the upper bound of model complexity and limits our ability to test architectures that may better capture nonlinear or distributed patterns. As larger datasets become available, future work may explore hybrid pipelines, transfer learning, and representation learning that leverage pre-trained models or multimodal embedding spaces to detect subtle cognitive states. Implementing standardized micro-phenomenological reporting could strengthen correlations with objective brain signatures (Czajko et al., 2024; Potash et al., 2024). Expanding into multimodal designs integrating fMRI with EEG, MEG, or diffusion imaging will be critical for resolving both spatial and temporal dynamics of ACAM-J (Potash et al., 2025; S. Singh et al., 2022; Treves et al., 2025).

## 5 Conclusion

This study provides evidence that advanced concentrative absorption states can be decoded from 7T fMRI ReHo patterns using machine-learning approaches grounded in first-person phenomenology. Distributed ReHo patterns distinguished ACAM-J from matched cognitively active control tasks, discriminated among ACAM-J states with performance that scaled with their phenomenological separation, and implicated prefrontal, cingulate, insular, parietal, and brainstem regions as the most consistent contributors to classification. The findings contribute to the understanding of the neural mechanisms underlying advanced meditation and demonstrate the feasibility of multivariate decoding in contemplative neuroscience, offering a basis for future multimodal and larger-scale studies of trained states of consciousness.

### Data and Code Availability

The code used in this study is publicly available at the GitHub repository: https://github.com/MIntelligence-Group/MRP. The human participant data underlying this study are not publicly available because they contain sensitive information and are subject to Institutional Review Board and institutional data-sharing restrictions. De-identified data may be made available to qualified researchers upon reasonable request, subject to Institutional Review Board approval and in accordance with the data-sharing policies of Massachusetts General Hospital. Requests for access should be directed to the Meditation Research Program Office, Department of Psychiatry, Massachusetts General Hospital, Boston, MA 02129, USA (meditationadministration@mgh.harvard.edu).

### Author Contributions

**Conceptualization**: Matthew D. Sacchet.

**Data curation**: Winson F. Z. Yang.

**Formal analysis**: Puneet Kumar, Alakhsimar Singh.

**Funding acquisition**: Matthew D. Sacchet.

**Investigation**: Winson F. Z. Yang.

**Methodology**: Puneet Kumar, Xiaobai Li.

**Software**: Puneet Kumar, Alakhsimar Singh.

**Project administration**: Matthew D. Sacchet.

**Supervision**: Matthew D. Sacchet, Xiaobai Li.

**Visualization**: Puneet Kumar, Alakhsimar Singh.

**Writing, original draft:** Puneet Kumar, Alakhsimar Singh.

**Writing, review and editing**: Winson F. Z. Yang, Xiaobai Li, Matthew D. Sacchet.

**Funding**

Dr. Sacchet and the Meditation Research Program are supported by the National Institute of Mental Health (Project Number R01MH125850), Dimension Giving Fund, Tan Teo Charitable Foundation, and individual donors.

**Declaration of Competing Interests**

All authors indicate no competing interests.

**Acknowledgements**

We are grateful for the advanced meditators whose meditative capacities made this study possible. We also respectfully acknowledge the Buddhist traditions that have preserved and transmitted *jhana* meditation for more than two thousand years. The present study builds upon this historical knowledge by examining these advanced absorptive practices with contemporary scientific methods. Our aim here is not to uncritically validate religious claims, but rather to investigate such practices, and their outcomes, as reproducible human capacities that promise to contribute to human well-being in the modern world.

**Use of Artificial Intelligence**

During manuscript preparation, the authors used ChatGPT (OpenAI) and Grammarly solely to improve language clarity. They subsequently reviewed and edited all AI-assisted text and take full responsibility for the content. No AI tools were used for data analysis, figure generation, or methodological design; all scientific content, results, and interpretations are the authors' own.